\documentclass{article} 
\usepackage{iclr2021_conference,times}

\usepackage{url}

\usepackage{makecell}
\usepackage{times}
\usepackage{epsfig}
\usepackage{graphicx}
\usepackage{amsmath}
\usepackage{amssymb}
\usepackage{booktabs}
\usepackage{multirow}
\usepackage{pifont}
\newcommand{\cmark}{\ding{51}}

\newcommand{\etal}{\textit{et al.}}

\usepackage[pagebackref=true,breaklinks=true,letterpaper=true,colorlinks,bookmarks=false]{hyperref}

\title{Revisiting 3D ResNets for Video Recognition}

\author{Xianzhi Du, Yeqing Li, Yin Cui, Rui Qian\thanks{Work done during an internship at Google.} , Jing Li\footnotemark[2], Irwan Bello\thanks{Equal advising.} \\
Google Research \\
\small{\texttt{\{xianzhi,yeqing,yincui,qianrui,jingli,ibello\}@google.com}} \\
}

\iclrfinalcopy 
\begin{document}
\maketitle
\begin{abstract}

A recent work from Bello~\etal~\cite{bello2021revisiting} shows that training and scaling strategies may be more significant than model architectures for visual recognition.
This short note studies effective training and scaling strategies for video recognition models.
We propose a simple scaling strategy for 3D ResNets, in combination with improved training strategies and minor architectural changes.
The resulting models, termed 3D ResNet-RS, attain competitive performance of 81.0\% on Kinetics-400 and 83.8\% on Kinetics-600 without pre-training.
When pre-trained on a large Web Video Text dataset, our best model achieves 83.5\% and 84.3\% on Kinetics-400 and Kinetics-600. 
Code is available at: \small{\url{https://github.com/tensorflow/models/tree/master/official}}.

\end{abstract}

\section{Introduction}
Recent studies have shown that vision models may be significantly improved through the combination of modern training and simple scaling strategies~\cite{bello2021revisiting, Du2021SimpleTS, Kolesnikov2020BigT}.
For example, Bello~\etal~\cite{bello2021revisiting} remark that the canonical ResNet-200~\cite{He2016ResNet} is improved from 79.0\% to 83.4\% (\textcolor{blue}{+4.4\%}) top-1 ImageNet accuracy through improved training methods and lightweight architectural changes.
The resulting ResNet-RS model is further scaled via model depth and image resolution, reaching a competitive accuracy of 84.4\% (\textcolor{blue}{+5.4\%}).

Inspired by ResNet-RS~\cite{bello2021revisiting}, our study focuses on effective training and scaling methods for video action recognition models. 
Using a 3D ResNet (R3D) model as the baseline, we evaluate the effect of \textbf{(1)} lightweight architectural changes, including the ResNet-D stem~\cite{He2019BagOT} and Squeeze-and-Excitation~\cite{Hu2018SqueezeandExcitationN} and \textbf{(2)} improved training methods such as data augmentation and regularization.
Further, we propose a simple scaling rule that simultaneously scales model depth and the temporal resolution of the input.

We evaluate the resulting models, referred to as 3D ResNet-RS (R3D-RS), on the popular Kinetics-400 and Kinetics-600 benchmarks. 
When trained from scratch, R3D-RS-50 model obtains a \textcolor{blue}{+3.8\%} top-1 accuracy improvement over the R3D-50 baseline. 
By further scaling to R3D-RS-200 with 48 input frames, our largest model achieves competitive top-1 accuracies of 81.0\% and 83.8\% on Kinetics-400 and Kinetics-600 respectively, on par with recent state-of-the-art models.
When pretrained on the Web Video Text dataset~\cite{stroud2020learning}, our R3D-RS-200 model is further improved by \textcolor{blue}{+2.5\%} on Kinetics-400 and \textcolor{blue}{+0.5\%} on Kinetics-600. 
Lastly, we evaluate R3D-RS in a self-supervised contrastive learning setup~\cite{qian2021spatiotemporal} where we demonstrate \textcolor{blue}{+1.4\%} top-1 Kinetics-400 accuracy improvement over the R3D baseline.

\section{Related Work}

\paragraph{Spatiotemporal networks for video recognition:}
A key difference between static images and videos is the additional temporal dimension. 
While some works model spatial and temporal features separately, it is more common to combine features through both the spatial and temporal dimensions by using 3D convolutions~\cite{taylor2010convolutional, tran2015learning, carreira2017quo} or self-attention~\cite{vaswani2017attention,Bello_2019_ICCV,ramachandran2019standalone,bello2021lambdanetworks,dosovitskiy2021image}

\paragraph{Data augmentation and regularization:}
When training data is insufficient or lacks diversity, regularization methods such as
data augmentation~\cite{Zhong2020RandomED, Devries2017ImprovedRO, Zhang2018mixupBE, Yun2019CutMixRS, Cubuk2019AutoAugmentLA, Cubuk2019RandAugmentPD}, 
dropout~\cite{dropout},
stochastic depth~\cite{stochasticdepth} or
dropblock~\cite{Ghiasi2018DropBlockAR}
are well known to improve generalization and mitigate overfitting.
Recent studies in image classification~\cite{bello2021revisiting} and object detection~\cite{Du2021SimpleTS} have shown that augmentation and regularization methods alone may significantly improve model accuracy without additional cost at inference.

\begin{figure}[t]
  \centering
  \includegraphics[width=0.6\columnwidth]{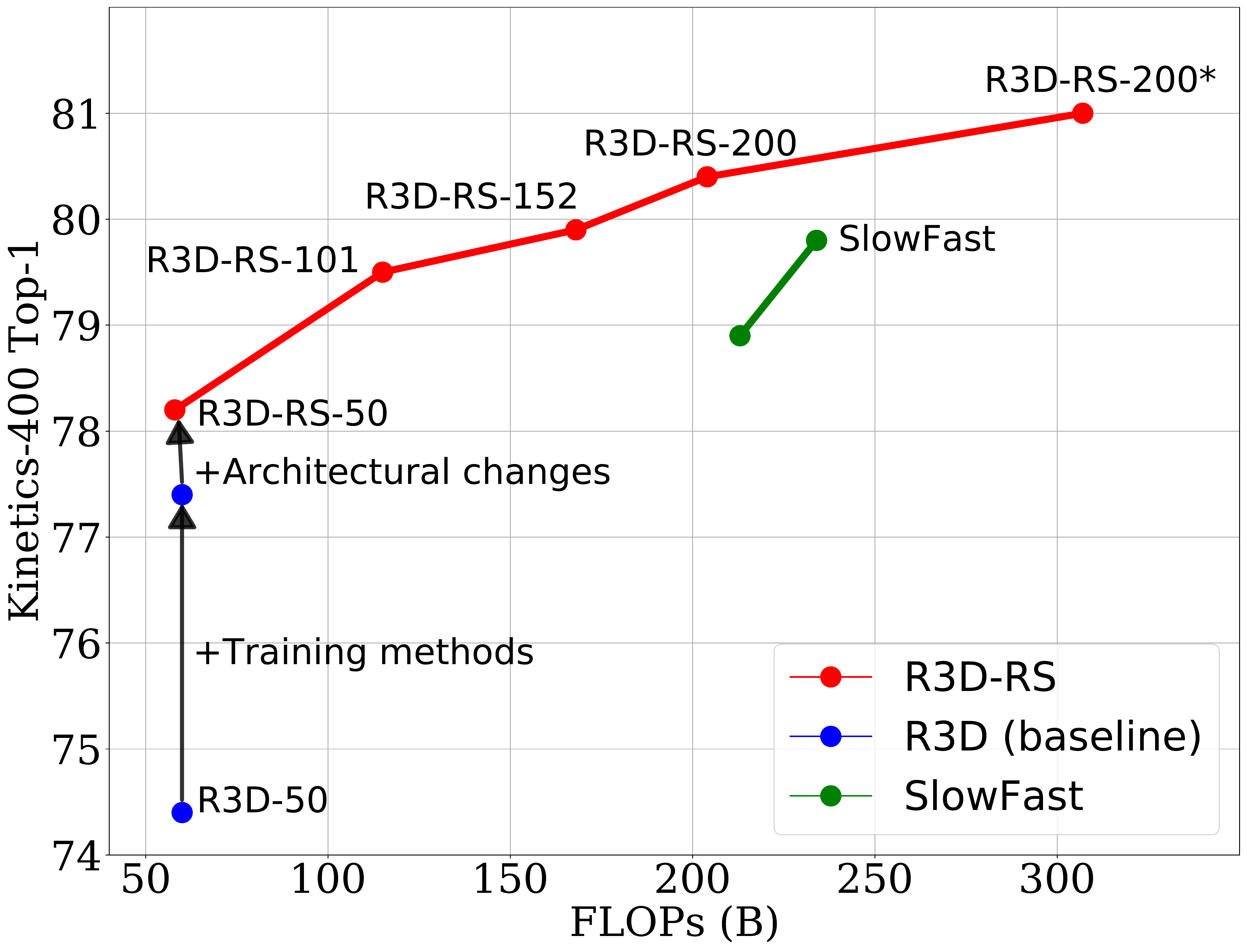}
\caption{\small{
Kinetics-400 top-1 accuracies of the R3D baseline, SlowFast networks and the proposed R3D-RS models.
The proposed training methods and architectural changes improve upon the R3D baseline by 3.8\% while slightly reducing computation.
The proposed simple scaling method further improves accuracy by 2.8\%.
R3D-RS-200$^\star$ scales up input frames and uses stronger augmentation. 
FLOPs are reported on a single inference view and all models use the same 30-view evaluation protocol. 
See Section~\ref{sec:exp} for experimental details. 
SlowFast accuracies are taken directly from~\cite{Feichtenhofer2019SlowFastNF} and may be improved through R3D-RS' training methods.
}}
\label{fig:page1_fig}
\vspace{-0mm}
\end{figure}

\section{Methodology}

\subsection{The 3D ResNet-RS architecture}
The full architecture specifications of our 3D ResNet-50 is presented in Table~\ref{tab:r50_spec}.
We start from R3D~\cite{qian2021spatiotemporal} as the baseline architecture and adopt the following lightweight architectural changes to improve model performance.

\paragraph{3D ResNet-D stem:} We adapt the ResNet-D~\cite{He2019BagOT} stem to 3D inputs by using three consecutive 3D convolutional layers.
The first convolutional layer employs a temporal kernel size of 5 while the remaining two convolutional layers employ a temporal kernel size of 1
Detailed layer specifications are shown in Table~\ref{tab:r50_spec} right. 

\paragraph{3D Squeeze-and-Excitation:}
We adapt Squeeze-and-Excite~\cite{Hu2018SqueezeandExcitationN} to spatio-temporal inputs by simply using a 3D global average pooling operation for the squeeze operation.
A SE ratio of 0.25 is applied in each 3D bottleneck block for all experiments. 

\paragraph{Self-gating:} We further plug in a self-gating module~\cite{xie2018rethinking} in each 3D bottleneck block after the SE module. 

\begin{table}[t]
\hspace{-5mm}
\begin{minipage}{.5\linewidth}
\centering
\scalebox{0.866}{
\begin{tabular}{c |  c}
  \toprule
  Block  & Operations  \\
  \midrule
  C2 Block &   \makecell{$1\times1^2, 64$ \\$1\times3^2, 64$\\$1\times1^2, 256$ \\$\text{SE-SG-SD}$  }  \\
  \midrule
  C3 Block &  \makecell{$1\times1^2, 128$ \\$1\times\underline{3}^2, 128$\\$1\times1^2, 512$ \\$\text{SE-SG-SD}$    }  \\
  \midrule
  C4 Block &   \makecell{$3\times1^2, 256$ \\$1\times\underline{3}^2, 256$\\$1\times1^2, 1024$ \\$\text{SE-SG-SD}$    }   \\
  \midrule
  C5 Block & \makecell{$3\times1^2, 512$ \\$1\times\underline{3}^2, 512$\\$1\times1^2, 2048 $\\$\text{SE-SG-SD}$ }  \\
  \bottomrule
\end{tabular}}
\end{minipage}
\hspace{-10mm}
\begin{minipage}{.5\linewidth}
\centering
\scalebox{0.9}{
\begin{tabular}{c | c  | c}
  \toprule
  Stage  & Output Size & Operations  \\
  \midrule
  Video clip & $32\times224^2$ & - \\
  \midrule
  Input & $16\times224^2$ & Downsample \\
  \midrule
  Stem & $8\times112^2$ &  \makecell{$\underline{5}\times\underline{3}^2, 32$ \\ $1\times3^2, 32$ \\ $1\times3^2, 64$ }\\
  \midrule
  Pool & $8\times56^2$ & $1\times\underline{3}^2$\\
  \midrule
  C2 & $8\times56^2$ & C2 Block$\times3$ \\
  \midrule
  C3 & $8\times28^2$ & C3 Block$\times4$ \\
  \midrule
  C4 & $8\times14^2$ & C4 Block$\times6$ \\
  \midrule
  C5 & $8\times7^2$ & C5 Block$\times3$ \\
  \midrule
  Head & $1\times1^2$ & \makecell{Ave Pool \\ DropOut \\ FC} \\
  \bottomrule
\end{tabular}}
\end{minipage}
\vspace{1mm}
\caption{\small{
\textbf{(Left)} Blocks used in the R3D-RS architecture.
 Kernel sizes are denoted as \{$\text{Temporal}\times\text{Spatial}^2, \text{Filters}$\} and the kernel is underlined when the spatial stride is 2 (1 otherwise).
SE: squeeze-and-excitation. 
SG: self-gating. 
SD: stochastic depth.
\textbf{(Right)} R3D-RS-50 architecture.
}}
\label{tab:r50_spec} 
\end{table}

\begin{table*}[t]
\centering
\begin{tabular}{l | c c cccccc | c | l }
\toprule
Model    &  LS & SD &  WD & 350-EP & SJ & SE & D-stem & RA & FLOPs (B) & Top-1  \\ 
\midrule
R3D-50 &    &  &  &  &   &  &  & & 60 & 74.4 \\ 
-- &  \cmark &  &  &  &  &  & & & 60 & 74.9  (\textcolor{blue}{+0.5}) \\ 
-- & \cmark & \cmark &  &  & & &  &  & 60 & 76.1 (\textcolor{blue}{+1.2}) \\ 
-- & \cmark & \cmark & \cmark & & & &  & & 60 & 76.3 (\textcolor{blue}{+0.2})  \\ 
-- & \cmark & \cmark & \cmark & \cmark & &&  &  & 60 & 76.4 (\textcolor{blue}{+0.1}) \\ 
-- & \cmark & \cmark & \cmark & \cmark & \cmark & & & & 60 & 77.4 (\textcolor{blue}{+1.0}) \\ 
-- & \cmark & \cmark & \cmark & \cmark & \cmark & \cmark  & &  & 60 & 77.9 (\textcolor{blue}{+0.5}) \\ 
R3D-RS-50 & \cmark & \cmark & \cmark & \cmark & \cmark & \cmark &  \cmark & & 58& 78.2 (\textcolor{blue}{+0.3}) \\
\midrule
R3D-RS-200 & \cmark & \cmark & \cmark & \cmark & \cmark & \cmark &  \cmark & & 307& 80.7 \\
R3D-RS-200 & \cmark & \cmark & \cmark & \cmark & \cmark & \cmark &  \cmark & \cmark & 307& 81.0 (\textcolor{blue}{+0.3})\\
\bottomrule
\end{tabular}
\vspace{1mm}
\caption{\small{
Additive study of the training methods and architectural changes used in this paper. LS: label smoothing. SD: stochastic depth. WD: reduced weight decay. 350-EP: increase training epochs to 350. SJ: scale jittering. SE: squeeze-and-excitation. D-stem: 3D ResNet-D stem. RA: 3D RandAugment. 
All models are trained from scratch on Kinetics-400 and evaluated using the 30-view protocol.}}
\label{tab:ablation_study} 
\end{table*}

\begin{figure}[h]
  \centering
  \includegraphics[width=0.6\columnwidth]{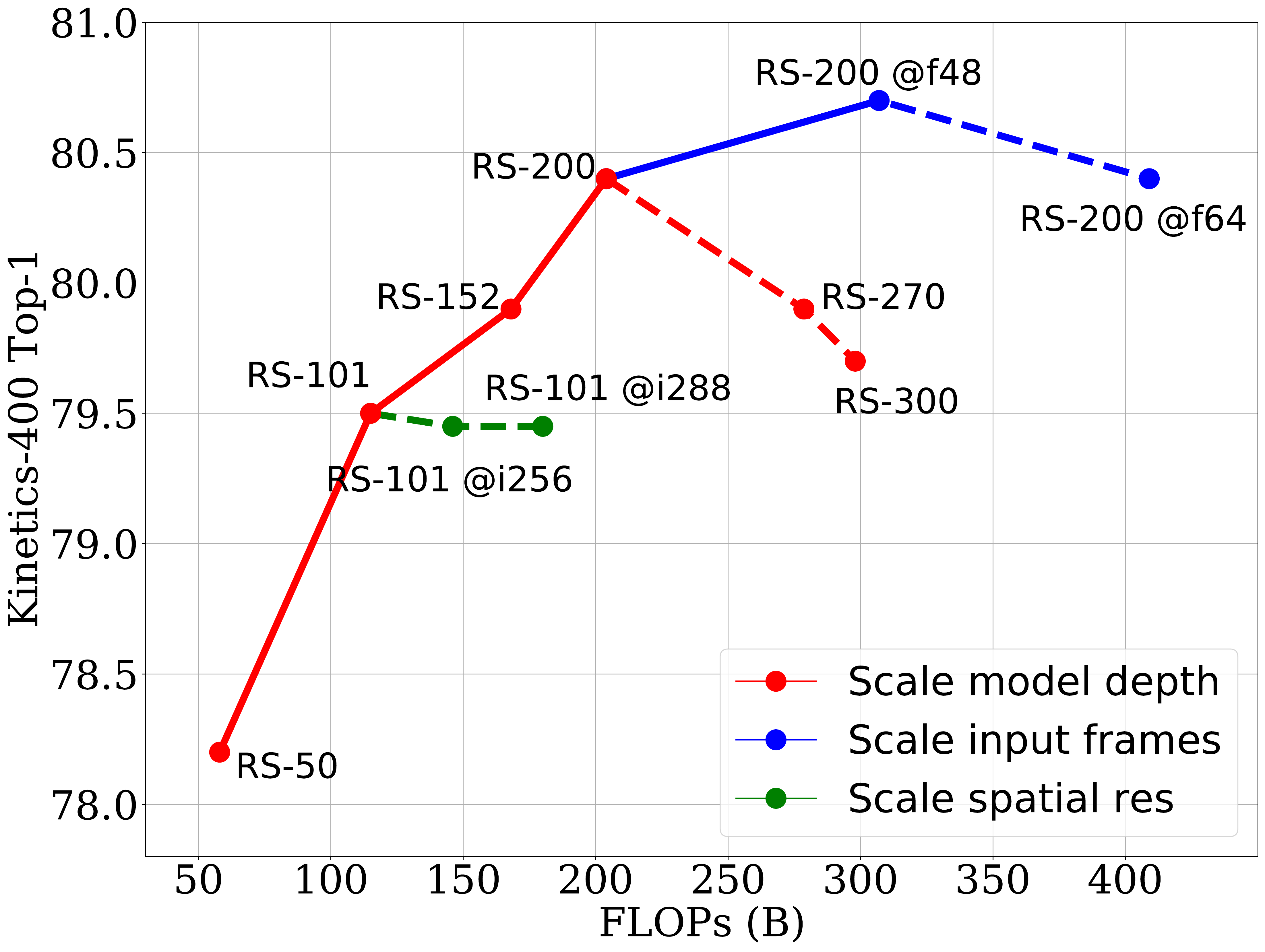}
\caption{\small{
Scaling 3D-ResNet-RS models via depth, spatial resolution or input frames.
Models are denoted as ``RS-depth''. 
Unless specified otherwise, models are trained from scratch on Kinetics-400 with input size 32$\times224^2$. ``@f48'' and ``@f64'' denote scaling input temporal frames to 48 and 64. 
``@i256'' and ``@i288'' denote scaling training input spatial resolution to 256 and 288. 
FLOPs are calculated on single view inference at 32$\times256^2$.
Details can be found in Sec.~\ref{sec:ablation_exp_1} and ~\ref{sec:ablation_exp_2}.}}
\label{fig:scaling}
\vspace{-0mm}
\end{figure}

\subsection{Improved Training Strategies}
\paragraph{Data augmentation:} We apply scaling jittering, random cropping and RandAugment~\cite{Cubuk2019RandAugmentPD} as our data augmentation methods. 
We apply the same augmentation strategy to all frames in a video clip, which we experimentally find to achieve good performance.

\paragraph{Regularization:} We train all models with weight decay, label smoothing and stochastic depth~\cite{stochasticdepth}. 
The stochastic depth~\cite{stochasticdepth} module is applied in each bottleneck block after the SE module and the self-gating module, as shown in Table~\ref{tab:r50_spec}. 

\subsection{Scaling Strategies for Video Recognition}
Similar to~\cite{bello2021revisiting}, we scale R3D-RS models by increasing model depth and input resolution. 
Fig~\ref{fig:scaling} presents our scaling analysis.
We found scaling the temporal dimension of the input to be more effective than scaling the  spatial dimensions.
We scale the number of frames of model input from 32 to 48 and keep the spatial resolution to $224\times224$ for training. 

\section{Experiments\label{sec:exp}}

\subsection{Experimental settings}
We experiment on the Kinetics-400~\cite{K400} and Kinetics-600~\cite{K600} benchmarks and report top-1 and top-5 accuracy as evaluation metrics.
\paragraph{Training from scratch:} Our main results are reported under the settings of training from \textit{scratch} on Kinetics-400 and Kinetics-600. We use SGD with a 0.9 momentum rate and a batch size of 1024 for 350 epochs on TPUv3 devices~\cite{jouppi2017tpu}. We apply a cosine decay learning rate schedule with an initial learning rate 0.8 and a linear learning rate warm-up that is applied over the first 5 epochs. 0.5 dropout~\cite{dropout} is applied for regularization.

\paragraph{Inference:} We adopt the 30 views protocol~\cite{Feichtenhofer2019SlowFastNF} to report inference results. 10 temporal clips are uniformly sampled along the temporal axis. For each clip, we use 3 256$\times$256 spatial crops. 

\setlength{\tabcolsep}{4pt}
\begin{table*}[t]
\centering
\begin{tabular}{l  | c | c | c c}
  \toprule
  Model  & FLOPs (B)$\times$Views & Params (M) & Top-1 & Top-5  \\
  \midrule
  R(2+1)D~\cite{r21d} & 75$\times$10$\times$1 & 61.8 &  72.0 & 90.0 \\
  ip-CSN-152~\cite{Tran2019VideoCW} & 109$\times$10$\times$3 & 32.8 &   77.8 & 92.8 \\
  SlowFast-16$\times$8, R101~\cite{Feichtenhofer2019SlowFastNF} & 213$\times$10$\times$3 & - &  78.9 & 93.5 \\
  SlowFast-16$\times$8, R101+NL~\cite{Feichtenhofer2019SlowFastNF} & 234$\times$10$\times$3 & 59.9 &  79.8 & 93.9 \\
  MViT-B, 32$\times$3~\cite{Fan2021MultiscaleVT} &170$\times$5$\times$1& 36.6 &  80.2 & 94.4 \\
  X3D-XXL~\cite{feichtenhofer2020x3d} & 48$\times$10$\times$3 & 20.3 &  80.4 & 94.6 \\
  MViT-B, 64$\times$3~\cite{Fan2021MultiscaleVT} & 455$\times$3$\times$3& 36.6 &  81.2 & 95.1 \\
  \midrule
  R3D-50 (baseline) & 60$\times$10$\times$3 & 38.1 &  74.4 & 91.0 \\
  R3D-RS-50  &  58$\times$10$\times$3&  48.2 & 78.2 & 93.7 \\
  R3D-RS-101 &  115$\times$10$\times$3 & 85.1 & 79.5 & 94.2 \\
  R3D-RS-152 & 168$\times$10$\times$3 & 114.9 & 79.9 & 94.3\\
  R3D-RS-200 & 204$\times$10$\times$3 & 121.6 & 80.4 & 94.4 \\
  R3D-RS-200 + RA (48$\uparrow$) &  307$\times$10$\times$3 & 121.6 & 81.0 & - \\
  \bottomrule
\end{tabular}
\vspace{1mm}
\caption{\small{
\textbf{Comparisons of models trained from \textit{scratch} on Kinetics-400.} 
Inference ``Views'' is presented in $\text{view}_{\text{temporal}} \times \text{view}_{\text{space}}$. 
``48$\uparrow$'': scaling up input \#frames from 32 to 48. RA: RandAugment for 3D video clip.}}
\label{tab:k400results_sv} 
\end{table*}

\setlength{\tabcolsep}{4pt}
\begin{table*}[b]
\centering
\begin{tabular}{l | c | c|   c}
  \toprule
  Model  & FLOPs$\times$Views & Params  & Top-1  \\
  \midrule
  SlowFast~\cite{Feichtenhofer2019SlowFastNF} &  213$\times$10$\times$3 & - & 81.1  \\
  SlowFast, +NL~\cite{Feichtenhofer2019SlowFastNF} &  234$\times$10$\times$3 & 59.9 & 81.8  \\
  X3D-XL~\cite{feichtenhofer2020x3d} & 48$\times$10$\times$3 & 20.3 & 81.9 \\
  MViT-B, 32$\times$3~\cite{Fan2021MultiscaleVT} &  170$\times$5$\times$1& 36.8 & 83.4 \\
  MViT-B-24, 32$\times$3~\cite{Fan2021MultiscaleVT} & 236$\times$5$\times$1& 52.9 &  83.8 \\
  \midrule
  R3D-RS-200 &  205$\times$10$\times$3 & 122.0 &  83.1 \\
  R3D-RS-200 (48$\uparrow$) &  307$\times$10$\times$3 & 122.0 & 83.8 \\
  \bottomrule
\end{tabular}
\vspace{1mm}
\caption{\small{
\textbf{Comparisons of models trained from \textit{scratch} on Kinetics-600.} Inference ``Views'' is presented in $\text{view}_{\text{temporal}} \times \text{view}_{\text{space}}$. ``48$\uparrow$'': scaling up input \#frames from 32 to 48. SlowFast networks adopt the 16$\times
$8 setting with a R101 backbone.}}
\label{tab:k600results_sv} 
\end{table*}

\subsection{Training from scratch results on Kinetics}

Table~\ref{tab:k400results_sv} presents our results of training from \textit{scratch} on the Kinetics-400 benchmark. Comparing to the R3D-50 baseline~\cite{bello2021revisiting}, the modern training methods and architectural changes introduced in R3D-RS-50 significantly improve the top-1 and top-5 accuracy by 3.8\% and 2.7\%, respectively. After scaling the model depth from R3D-RS-50 to R3D-RS-200, the top-1 and top-5 accuracy are improved by 2.2\% and 0.7\%, respectively. The top-1 accuracy is further improved by 0.3\% by scaling input frames from 32 to 48. Lastly, adopting the 3D RandAugment strategy improves the top-1 accuracy by another 0.3\%.

We further evaluate our best R3D-RS-200 models on Kinetics-600 and report the results in Table~\ref{tab:k600results_sv}.

\subsection{Ablation of the training methods and architectural changes}\label{sec:ablation_exp_1}
We present detailed ablation studies of our training methods and architectural changes in Table~\ref{tab:ablation_study} on Kinetics-400 and report top-1 accuracy. Starting from the R3D-50 baseline, we gradually apply 0.1 label smoothing (\textcolor{blue}{+0.5\%}), stochastic depth with an initial drop rate of 0.2 (\textcolor{blue}{+1.2\%}), 4e-5 weight decay (\textcolor{blue}{+0.2\%}), prolonging training epochs to 350 (\textcolor{blue}{+0.1\%}), scale jittering augmentation (\textcolor{blue}{+1.0\%}), squeeze-and-excitation module with 0.25 rate (\textcolor{blue}{+0.5\%}), 3D ResNet-D stem (\textcolor{blue}{+0.3\%}) and RandAugment (\textcolor{blue}{+0.3\%}). The 3D ResNet-D stem also slightly reduces model FLOPs from 60B to 58B.

\subsection{Effectiveness of the simple scaling rule}\label{sec:ablation_exp_2}
The effectiveness of our simple scaling method is studied in Fig.~\ref{fig:scaling}. Scaling from R3D-RS-50 to R3D-RS-200 improves accuracy by \textcolor{blue}{+2.2\%}. However, further depth scaling from R3D-RS-200 to R3D-RS-270 (\textcolor{red}{-0.5\%}) or R3D-RS-300 (\textcolor{red}{-0.7\%}) hurt performance. Scaling input frames from 32 to 48 improves another \textcolor{blue}{+0.3\%}. We further studied scaling training (inference) spatial resolution from 224 (256) to 256 (288) or larger but the model accuracy drops. All models are trained from scratch on Kinetics-400 and top-1 accuracy are reported.

\subsection{Improvements from pretraining}
Table~\ref{tab:wvt_pretrain_results_sv} shows the results of using large scale pre-train dataset Web Videos and Text (WVT)~\cite{stroud2020learning}.
The WVT dataset contains 70 million video clips collected by using labels of the Kinetics-700 data set as query string in YouTube.
For more details of the data set please refer to~\cite{stroud2020learning}. We pretrain our R3D-RS-200 under the video classification 
task (i.e. input video clips to predict the action labels). We used the same input size (48$\times$224$\times$224) as previous experiments.
We pretrain the model on WVT for 4 epochs using learning rate 1.6, SGD with momentum 0.9 on TPUv3 128 cores.
The pretrained model is finetuned on Kinetics-\{400,600\} for 10000 steps with learning rate 0.05.
We have observed various level of improvements of the final model quality on Kinetics-400 (\textcolor{blue}{+2.5\%}) and Kinetics-600 (\textcolor{blue}{+0.5\%}).

\begin{table}[h]
\centering
\begin{tabular}{l | c | c | c |c}
\toprule
Model            & Pretrain & Dataset & FLOPs (B)$\times$Views & Top-1  \\ 
\midrule
VTN~\cite{Neimark2021VideoTN} & ImageNet-21K& Kinetics-400 & 4218$\times$1$\times$1 & 78.6 \\
TimeSformer-L~\cite{Bertasius2021IsSA} & ImageNet-21K& Kinetics-400 & 2380$\times$1$\times$3 & 80.7 \\
ViViT-L/16$\times$2 FE~\cite{Arnab2021ViViTAV} & ImageNet-21K &  Kinetics-400 & 3980$\times$1$\times$3 & 81.7 \\ 
Swin-L~\cite{liu2021videoswin} & ImageNet-21K&  Kinetics-400 & 604$\times$4$\times$3 & 83.1 \\
Swin-L, 384~\cite{liu2021videoswin} & ImageNet-21K&  Kinetics-400 & 2107$\times$10$\times$5 & 84.9 \\
ViViT-H/16$\times$2~\cite{Arnab2021ViViTAV} & JFT &  Kinetics-400 & 3981$\times$4$\times$3 & 84.9 \\
\midrule
R3D-RS-200 (48$\uparrow$) & WVT      & Kinetics-400 & 307$\times$10$\times$3 & 83.5 \\ 
\midrule
\midrule
TimeSformer-L~\cite{Bertasius2021IsSA} & ImageNet-21K& Kinetics-600 & 2380$\times$1$\times$3 & 82.2 \\
Swin-B~\cite{liu2021videoswin} & ImageNet-21K&  Kinetics-600 & 282$\times$4$\times$3 & 84.0 \\
ViViT-H/16$\times$2~\cite{Arnab2021ViViTAV}& JFT &  Kinetics-600 & 3891$\times$4$\times$3 & 85.8 \\
Swin-L, 384~\cite{liu2021videoswin} & ImageNet-21K&  Kinetics-600 & 2107$\times$10$\times$5 & 86.1 \\
\midrule
R3D-RS-200 (48$\uparrow$) & WVT      & Kinetics-600 & 307$\times$10$\times$3 &84.3 \\ 
\bottomrule
\end{tabular}
\vspace{1mm}
\caption{\small{
Kinetics-400/600 result comparisons of our R3D-RS models and other models when pretrained on large-scale datasets.}}
\label{tab:wvt_pretrain_results_sv} 
\end{table}

\section{Experiments of Unsupervised Learning}
We further test the model scaling strategy in self-supervised manner. For simplicity, we use the algorithm of CVRL~\cite{qian2021spatiotemporal} which learns spatiotemporal invariance features from short video clips. We pretrain the models for 800 epochs to ensure convergence. Pretraining batchsize is set to 2048 and we adopt lars optimizer with an initial learning rate of 0.64 to stablize training. For linear evaluation, we simply adopt SGD with initial learning rate of 64 and train for 100 epochs with batchsize of 1024. We report the top-1 accuracy on Kinetics-400 in Table~\ref{tab:k400_ssl}. Aligning with observation in Fig.~\ref{fig:scaling}, we find the performance saturating phenomenon as the model depth increase in self-supervised learning as well. 

\setlength{\tabcolsep}{4pt}
\begin{table}[h!]
\centering
\begin{tabular}{c | c| c|c|c|c}
  \toprule
  Model  & 
  R3D-RS-50 & R3D-RS-101 & R3D-RS-152 & R3D-RS-200 & R3D-RS-270\\
  \midrule
 Top-1 & 66.1 & 67.1 (\textcolor{blue}{+1.0}) &67.1 (\textcolor{blue}{+0.0}) & 67.3 (\textcolor{blue}{+0.2}) & 67.5 (\textcolor{blue}{+0.2}) \\
  \bottomrule
\end{tabular}
\vspace{1mm}
\caption{\small{
Linear evaluation results of R3D-RS models based on self-supervised pretraining on Kinetics-400.
}}
\label{tab:k400_ssl} 
\end{table}

\section{Conclusion}
In this work, we revisit 3D ResNets in the light of the modern techniques that commonly applied in image classification
The proposed training and scaling strategies, along with lightweight architectural changes, significantly improve 3D ResNet models when trained from scratch on the popular Kinetics-400 and Kinetics-600 benchmarks.
We believe our methods can benefit more models for action recognition and boarder video applications.

\clearpage
{\small
\bibliographystyle{ieee_fullname}
\bibliography{egbib}
}

\end{document}